\documentclass[10pt,twocolumn,letterpaper]{article}

\usepackage[final,pagenumbers]{cvpr}  

\usepackage[utf8]{inputenc}
\usepackage{multirow}
\usepackage[ruled]{algorithm}
\usepackage{algpseudocode}
\usepackage{microtype} 

\definecolor{cvprblue}{rgb}{0.21,0.49,0.74}
\usepackage[breaklinks,colorlinks,allcolors=cvprblue]{hyperref}

\def\confName{CVPR}
\def\confYear{2027}

\title{TIDE: Token-Level Credit Assignment via Entropy-Gated Bounded Deviation}

\author{Qiaoman Zhang$^{1*}$\quad Shunhua Min$^{2*}$\quad Xiang He$^{3*}$\quad Zhenghan Zhang$^{1}$\quad Yuting Zhang$^{1}$\\
Guangxian Li$^{1}$\quad Yunzhu Kuang$^{1}$\quad Rui Jiang$^{1}$\quad Jingyan Zhang$^{1}$\quad Chaoyue Li$^{4}$\quad Jiayu Ding$^{1\dagger}$\\[2pt]
{\small$^{1}$Peking University\quad $^{2}$Shenzhen University\quad $^{3}$Xi'an Jiaotong University}\\
{\small$^{4}$Huazhong University of Science and Technology}\\[2pt]
{\small$^{*}$Equal contribution.\quad $^{\dagger}$Corresponding author.}
}

\begin{document}

\maketitle

\begin{abstract}
Reinforcement learning (RL) is central for aligning large language models (LLMs) with complex reasoning, yet the leading critic-free algorithm, Group Relative Policy Optimization (GRPO), broadcasts a single sequence-level advantage uniformly across long Chain-of-Thought (CoT) generations and constrains exploration with an unbounded Kullback-Leibler (KL) penalty, which dilutes token-level supervision and induces gradient instability and mode-seeking conservatism.
We propose TIDE (Token-level Importance via Deviation and Entropy), a critic-free RL framework that reinterprets distribution deviation as a \emph{guiding signal} rather than a rigid penalty.
TIDE replaces the volatile reverse KL divergence with the bounded Hellinger distance, safely quantifying token-level exploration without risk of gradient explosion.
To separate genuine reasoning breakthroughs from confident hallucinations, an entropy gating mechanism scales this deviation by the policy's predictive uncertainty, and the gated scores redistribute the sequence-level advantage across tokens.
The resulting updates amplify exploratory steps while suppressing low-entropy deviations, without a value network and with negligible overhead.
Across in-domain, held-out, and out-of-domain benchmarks, TIDE outperforms strong critic-free baselines: on the Qwen2.5-32B backbone it reaches 60.0\% and 46.0\% Avg@32 accuracy on AIME 2024 and AIME 2025, respectively, above competitive baselines such as DAPO.
Token-level attribution, causal perturbation, and exploration analyses further support the interpretation that the entropy-gated deviation identifies pivotal reasoning steps rather than merely perturbing optimization dynamics.
\end{abstract}

\section{Introduction}
\label{sec:intro}

Aligning large language models (LLMs) with reinforcement learning (RL) has fundamentally advanced their capacity for complex problem solving and logical reasoning \cite{DBLP:conf/nips/Ouyang0JAWMZASR22, guo2025deepseek}.
Among recent paradigms, Group Relative Policy Optimization (GRPO) \cite{shao2024deepseekmath} is particularly successful at eliciting long-horizon Chain-of-Thought (CoT) behavior without an auxiliary value network: by optimizing the policy relative to a group of sampled responses, it reduces memory overhead while remaining competitive on mathematical and algorithmic reasoning \cite{guo2025deepseek}.
In this setting, models learn to self-correct and explore extended logical trajectories purely through scalar rewards from verifiers or reward models.

Despite these advantages, current critic-free RL frameworks suffer from a severe \emph{credit assignment} problem \cite{sutton1998reinforcement, DBLP:conf/iclr/RamamurthyABHSB23}.
In standard GRPO, a single scalar advantage is broadcast uniformly across an entire generated sequence, which often spans thousands of tokens.
This coarse assignment forces the model to treat pivotal reasoning breakthroughs and redundant transitional syntax with equal importance, diluting the learning signal.
Compounding this inefficiency, the standard Kullback-Leibler (KL) penalty \cite{kullback1951information, ziegler2019fine} constrains the policy to a reference model $\pi_{\mathrm{ref}}$: the unbounded reverse KL enforces extreme mode-seeking conservatism, harshly penalizing any deviation from $\pi_{\mathrm{ref}}$.
When the model explores novel, low-probability reasoning trajectories, this unbounded penalty causes severe gradient instability, disincentivizing innovative solutions and trapping reasoning within the safe yet restricted bounds of the pre-trained distribution \cite{DBLP:conf/nips/KorbakEKD22}.

\begin{figure*}[t]
    \centering
    \includegraphics[width=\linewidth]{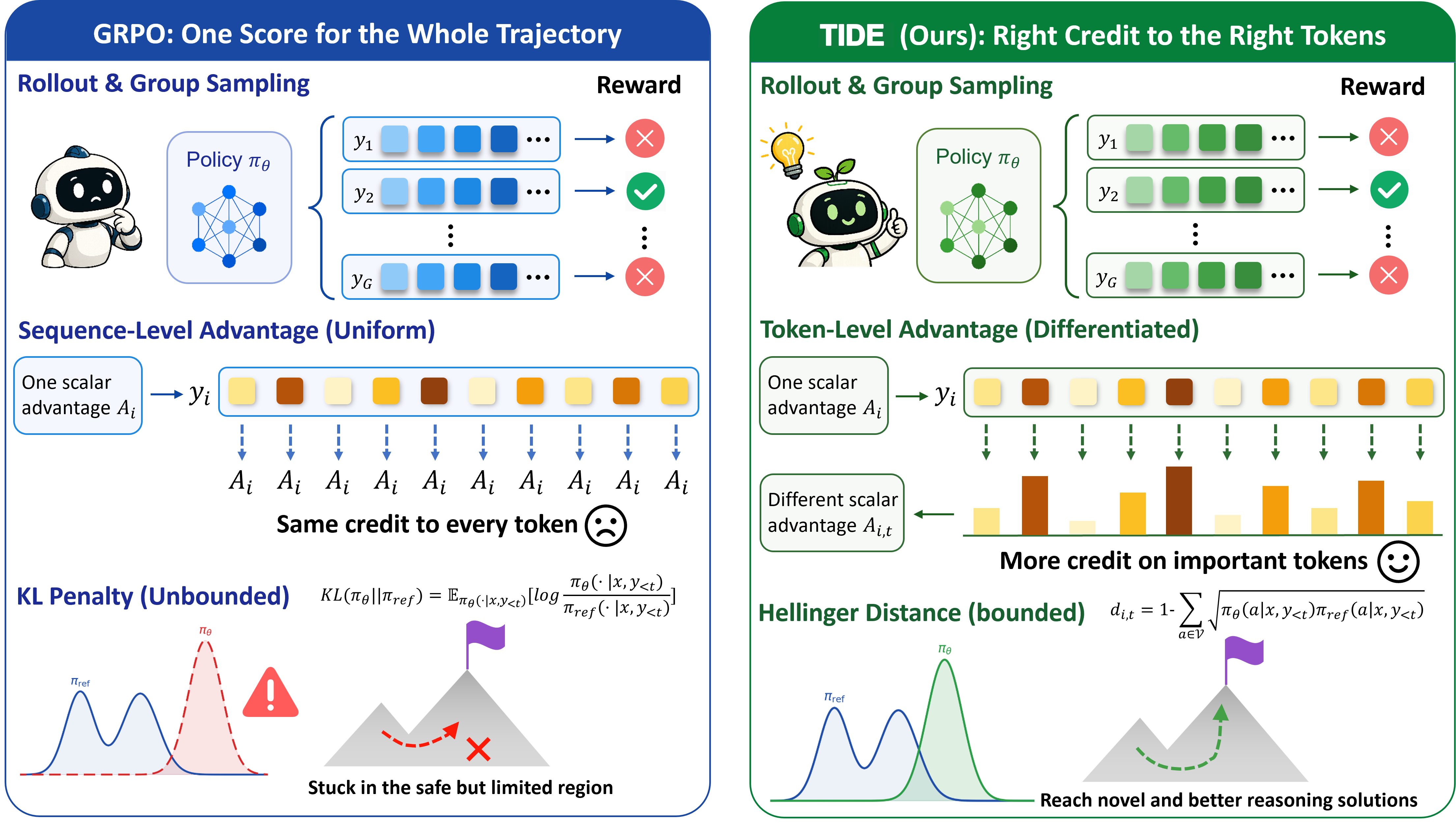}
    \caption{\textbf{Conceptual comparison between standard GRPO and TIDE.} While GRPO uniformly broadcasts a coarse, sequence-level advantage and imposes an unbounded reverse KL penalty that stifles exploration, TIDE dynamically reallocates advantages to individual tokens according to an entropy-gated, bounded deviation signal.}
    \label{fig:comparison}
\end{figure*}

To overcome these bottlenecks, we propose \emph{TIDE} (Token-level Importance via Deviation and Entropy), an RL framework that reallocates coarse sequence-level advantages into fine-grained, token-level update signals.
TIDE is built on a new perspective: deviation from the reference distribution should not be uniformly penalized as an error, but interpreted as a \emph{guiding signal for exploration}.
When a sequence yields a positive reward, the tokens that diverge most from the reference model are often exactly the cognitive leaps responsible for the success.
To quantify this deviation safely, TIDE replaces the reverse KL divergence with the bounded Hellinger distance \cite{beran1977minimum, amari2016information, DBLP:conf/iclr/WangJYLC24}, which smoothly encourages exploratory diversity without the risk of gradient explosion.

Relying solely on distribution deviation, however, risks rewarding ``fake innovations'': a large deviation may reflect spurious or hallucinated tokens rather than a meaningful reasoning step.
When the model assigns high confidence to an incorrect, out-of-distribution token, the resulting deviation can be large despite lacking semantic validity.
Such confident hallucinations would be incorrectly reinforced by deviation-based signals alone.
TIDE therefore introduces an entropy gating mechanism that scales the Hellinger distance by the model's normalized predictive uncertainty (entropy), distinguishing deliberate, high-uncertainty exploration from confident, low-entropy hallucinations.
The gated scores then dynamically redistribute the sequence-level advantage across tokens: critical exploratory steps receive amplified gradients, while standard syntactic tokens are naturally discounted.

We evaluate TIDE on the AIME 2024 and AIME 2025 mathematical reasoning benchmarks.
Across model scales, TIDE outperforms competitive critic-free alignment methods, achieving 60.0\% Avg@32 accuracy on AIME 2024 and 46.0\% on AIME 2025 with a Qwen2.5-32B backbone, above strong baselines including DAPO \cite{yu2025dapo}.
Beyond outcome accuracy, our computational profiling shows that TIDE preserves the efficiency of standard GRPO: it obtains process-level supervisory signals with only 3.6\% time overhead and a memory footprint comparable to vanilla GRPO, since it reuses logits that GRPO already materializes for its KL term.

In summary, our contributions are three-fold:
\textbf{(i)} We analyze the limitations of uniform sequence-level updates and the unbounded reverse KL divergence in long-horizon tasks, showing how they induce mode-seeking conservatism and gradient instability.
\textbf{(ii)} We propose a critic-free RL framework that reinterprets distribution deviation as a guiding signal: by combining the bounded Hellinger distance with entropy gating, TIDE safely incentivizes critical exploratory steps while filtering out confident hallucinations.
\textbf{(iii)} We validate TIDE across model scales (7B and 32B) on in-domain, held-out math, and out-of-domain science and code benchmarks with multi-seed confidence intervals and a paired significance test, and further validate the mechanism itself through token-level attribution, causal perturbation, and exploration analyses.

\section{Related Work}
\label{sec:related}

\paragraph{Reinforcement learning for LLM reasoning.}
RL has become the de facto standard for aligning LLMs with human intent and enhancing complex reasoning \cite{guo2025deepseek, DBLP:conf/nips/Ouyang0JAWMZASR22}.
While Proximal Policy Optimization (PPO) \cite{schulman2017proximal} initially dominated RLHF, its reliance on an auxiliary value network imposes prohibitive memory and computational overheads as parameters scale.
This bottleneck has driven a paradigm shift towards critic-free optimization \cite{feng2025group, guo2025deepseek, jin2025videocurlvideocurriculumreinforcement, jin2026himachierarchicalmacromicrolearning, DBLP:conf/nips/RafailovSMMEF23, yu2025dapo, zheng2025group}.
Among these, GRPO \cite{shao2024deepseekmath} is exceptionally effective at eliciting mathematical and logical reasoning, empowering models to perform extensive CoT generation \cite{zhou2026demystifying} by estimating the baseline through group-wise reward normalization rather than a parameterized critic.
However, GRPO applies a uniform, sequence-level advantage to all tokens in a response \cite{li2026distribution}, treating every decoding step equally; in long-horizon reasoning this inevitably obscures the learning signal and slows convergence.

\paragraph{Fine-grained credit assignment.}
The credit assignment problem is a fundamental challenge in RL, uniquely exacerbated in autoregressive LLMs \cite{DBLP:conf/iclr/RamamurthyABHSB23} where action spaces span tens of thousands of tokens.
To provide more granular supervision, recent work has heavily investigated Process Reward Models (PRMs) \cite{DBLP:conf/iclr/LightmanKBEBLLS24, uesato2022solving, zhang2025groundedprm, she2025r, yin2025dynamic, liu2025adaptivestep} and implicit or Monte-Carlo process rewards \cite{cui2025prime, kazemnejad2024vineppo}, which assign scalar rewards to intermediate reasoning steps, unlike Outcome Reward Models \cite{cobbe2021training, chen2025rm, zhang2025linking, dou2025pre, ye2025beyond} that evaluate only the final answer.
Despite their effectiveness, training robust PRMs requires exhaustive human-annotated step-level data or additional rollouts.
Closer to our setting, several works redistribute sequence-level rewards using attention weights or heuristic decay mechanisms \cite{DBLP:conf/icml/ZengLMYZW24, DBLP:conf/icml/LiXZL00L24}, and recent studies show that policy entropy itself marks decision-critical ``forking'' tokens \cite{wang2025forking, cui2025entropy}.
TIDE differs in that the reweighting signal is the joint product of entropy and the bounded Hellinger deviation from the reference policy: our ablations show that neither signal alone recovers the full gain, indicating that the deviation carries information beyond entropy.
TIDE bypasses external PRMs and parameterized critics entirely: by extracting inherent probabilistic deviations between the current and reference policies, it establishes a self-contained mechanism for token-level credit assignment.

\section{The TIDE Method}
\label{sec:method}

\subsection{Preliminaries and Motivation}
\label{sec:prelim}
In standard GRPO for CoT alignment, given a prompt $x$, the policy $\pi_\theta$ generates a group of candidate responses $\{y_1, \dots, y_G\}$, which are evaluated to yield scalar rewards $\{r_1, \dots, r_G\}$.
The sequence-level advantage is computed by group-wise Z-score normalization:
\begin{equation}
  A_i = (r_i - \hat{\mu}_r) \,/\, (\hat{\sigma}_r + \varepsilon),
  \label{eq:grpo}
\end{equation}
where $\hat{\mu}_r$ and $\hat{\sigma}_r$ are the group mean and standard deviation of the rewards.
This paradigm suffers from two critical limitations.
First, $A_i$ is distributed uniformly across all tokens of a potentially long sequence, so the model cannot distinguish pivotal reasoning breakthroughs from redundant transitional phrases.
Second, the objective incorporates an unbounded KL penalty that constrains the policy to a reference model $\pi_{\mathrm{ref}}$: when the policy explores trajectories where $\pi_{\mathrm{ref}}$ assigns near-zero probability, the penalty can explode, leading to severe gradient instability.
TIDE addresses both issues by reinterpreting distribution deviation not as a penalty but as a guiding signal for critic-free, fine-grained credit assignment.

\begin{figure*}[t]
    \centering
    \includegraphics[width=\linewidth]{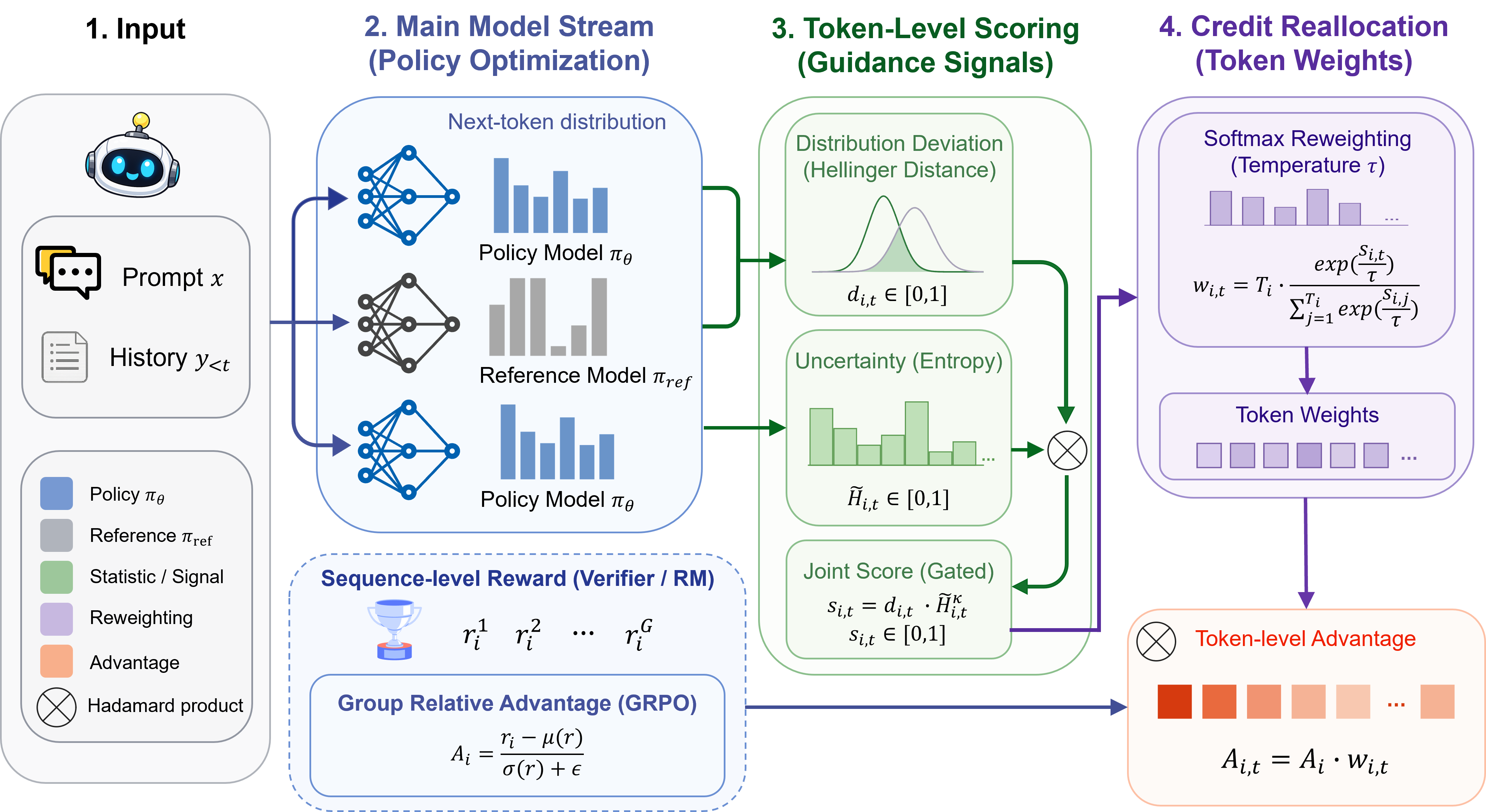}
    \caption{\textbf{The computational pipeline of TIDE.} The bounded Hellinger deviation between the current and reference policies is gated by the policy entropy, converted into unit-mean redistribution weights, and used to reallocate the sequence-level advantage to individual tokens.}
    \label{fig:pipeline}
\end{figure*}

\subsection{Advantage Redistribution via Entropy-Gated Deviation}
\label{sec:redistribution}

Instead of treating the distribution constraint as an extrinsic penalty (as the KL term in PPO or GRPO), TIDE internalizes the regularization by reinterpreting distribution deviation as a token-level guidance signal, transforming the coarse advantage $A_i$ into a fine-grained credit assignment via a redistribution weight $w_{i,t}$.

A primary challenge in RLHF \cite{DBLP:conf/nips/Ouyang0JAWMZASR22, ziegler2019fine} is the numerical instability of the unbounded KL divergence, especially when $\pi_\theta$ explores regions where $\pi_{\mathrm{ref}}$ assigns near-zero probability.
We instead employ the Hellinger distance to quantify the deviation at each step $t$:
\begin{equation}
\begin{aligned}
d_{i,t} &= H^2\big(\pi_\theta(\cdot|x, y_{<t}),\, \pi_{\mathrm{ref}}(\cdot|x, y_{<t})\big) \\
&= 1 - \sum_{a \in \mathcal{V}} \sqrt{\pi_\theta(a|x, y_{<t})\,\pi_{\mathrm{ref}}(a|x, y_{<t})},
\end{aligned}
\label{eq:hdistance}
\end{equation}
where $d_{i,t} \in [0, 1]$.
Its intrinsic boundedness ensures the guidance signal remains bounded even during aggressive exploration, preventing the ``penalty explosion'' of KL-based objectives.
This is a deliberate design choice: TIDE uses the deviation as a \emph{multiplicative guidance signal} on the advantage rather than an \emph{additive penalty} in the loss.
An additive Hellinger penalty, like KL, would still pull the policy back toward $\pi_{\mathrm{ref}}$ and dampen exploration—the mode-seeking behavior we aim to avoid.
Reweighting instead redirects existing credit toward exploratory tokens without any restoring force toward the reference.

However, relying solely on distribution deviation makes the algorithm susceptible to ``fake innovations,'' such as the model confidently hallucinating a rare, out-of-distribution word \cite{wang2025arbitrary, sheng2025espo}.
TIDE therefore introduces an entropy gate that scales the deviation by the model's normalized predictive uncertainty.
Let $\tilde{\mathcal{H}}_{i,t} = \mathcal{H}\big(\pi_\theta(\cdot|x, y_{<t})\big) / \log|\mathcal{V}| \in [0,1]$ denote the Shannon entropy of the current policy at step $t$, normalized by the maximum entropy $\log|\mathcal{V}|$ over the vocabulary $\mathcal{V}$, and let $\kappa \ge 0$ control the gate strength.
We define the entropy-gated deviation score as
\begin{equation}
s_{i,t} = d_{i,t} \cdot \tilde{\mathcal{H}}_{i,t}^{\kappa}
\label{eq:score}
\end{equation}
for positive-advantage trajectories, so that $s_{i,t} \in [0,1]$ is large only when a token is \emph{both} highly divergent from the reference policy \emph{and} generated under high uncertainty, thereby distinguishing deliberate exploration from confident hallucinations.
The redistribution weight $w_{i,t}$ is obtained via a temperature-scaled softmax over $s_{i,t}$:
\begin{equation}
w_{i,t} = T_i \cdot \frac{\exp(s_{i,t} / \tau)}{\sum_{j=1}^{T_i} \exp(s_{i,j} / \tau)},
\label{eq:weight}
\end{equation}
where $\tau$ is a temperature and $T_i$ the sequence length.
Two properties follow directly.
First, the factor $T_i$ enforces the unit-mean property $\frac{1}{T_i} \sum_t w_{i,t} = 1$, preserving the overall gradient magnitude of the group.
Second, since $s_{i,t}\in[0,1]$, the per-token weight relative to the trajectory mean is bounded as $w_{i,t}\in[e^{-1/\tau},\, e^{1/\tau}]$; at the operating value $\tau=0.5$ this bounds the amplification-to-suppression ratio within any trajectory by $e^{1/\tau}\approx 7.4$, so the reweighting is aggressive but strictly controlled.

\paragraph{Gradient stability analysis.}
The stability benefit of TIDE stems from two design choices: \emph{removing the additive KL penalty} and the \emph{stop-gradient} treatment of the weights—rather than from any gradient property of the Hellinger distance itself, which, being detached, exercises no gradient path.
In KL-penalized RL, the penalty contributes $\beta \nabla_\theta \log \pi_\theta \,(\log \frac{\pi_\theta}{\pi_{\mathrm{ref}}} + 1)$ to the gradient; for the full-distribution KL this term becomes unbounded as $\pi_{\mathrm{ref}} \to 0$, and even with the bounded sampled-token $k_3$ estimator commonly used in practice, the penalty still exerts a restoring force toward $\pi_{\mathrm{ref}}$ that dampens exploration.
TIDE removes the penalty entirely, and an implementation detail makes our analysis exact: $w_{i,t}$ is treated as a \emph{stop-gradient} (detached) constant—the deviation $d_{i,t}$ and entropy $\tilde{\mathcal{H}}_{i,t}$ forming $w_{i,t}$ propagate no gradients to $\theta$.
The gradient of the TIDE objective (Eq.~\ref{eq:loss}) is therefore the exact identity
\begin{equation}
\nabla_\theta \mathcal{L}^{\mathrm{TIDE}} = \mathbb{E} \left[ w_{i,t} \cdot A_i \cdot \nabla_\theta \log \pi_\theta \right],
\label{eq:tide-grad}
\end{equation}
rather than an approximation: the divergence signal enters solely as a bounded multiplicative factor within the detached $w_{i,t}$, whose ratio to the trajectory mean is capped by $e^{1/\tau}$ as shown above.
TIDE therefore acts as a \emph{preconditioned} policy gradient: a state-dependently reweighted, unit-mean rescaling of the vanilla gradient whose direction remains ascent-aligned with the original objective under mild assumptions.
The complete boundedness analysis and the formal sign-consistency proposition are provided in Appendix~A.

\subsection{Fine-Grained Credit Reallocation and Objective}
\label{sec:objective}

The final step reallocates the sequence-level advantage $A_i$ to individual tokens using $w_{i,t}$ (Eq.~\ref{eq:weight}).
The unit-mean property $\frac{1}{T_i}\sum_t w_{i,t}=1$ preserves the magnitude of the optimization signal and prevents gradient vanishing across long sequences.
The coarse advantage is redistributed to yield the fine-grained local advantage $A_{i,t} = A_i \cdot w_{i,t}$, so that pivotal exploratory tokens receive amplified update signals while grammatical tokens receive diminished gradients.
Because the reference distribution is internalized within the localized advantage, TIDE removes the token-level KL penalty from the loss entirely.
The final objective is
\begin{equation}
\begin{aligned}
\mathcal{L}^{\mathrm{TIDE}}(\theta) = \frac{1}{G} \sum_{i=1}^{G} \frac{1}{T_i} \sum_{t=1}^{T_i} \min\big(&\rho_{i,t}A_{i,t}, \\
&\mathrm{clip}(\rho_{i,t}, 1{-}\epsilon_c, 1{+}\epsilon_c)\,A_{i,t}\big)
\end{aligned}
\label{eq:loss}
\end{equation}
where $\rho_{i,t} = \frac{\pi_\theta(y_{i,t} \mid x, y_{i,<t})}{\pi_{\theta_{\mathrm{old}}}(y_{i,t} \mid x, y_{i,<t})}$ is the standard importance sampling ratio between the current policy $\pi_\theta$ and the behavior policy $\pi_{\theta_{\mathrm{old}}}$.
By exploiting its own probabilistic dynamics, TIDE achieves stable, critic-free token-level credit assignment for long-sequence generation.

\paragraph{Handling of negative advantages.}
A subtle asymmetry arises from the sign of $A_i$.
For positive-reward trajectories ($A_i>0$), amplifying high-deviation tokens correctly reinforces the exploratory steps responsible for success.
For negative-reward trajectories ($A_i<0$), amplifying the same tokens would instead \emph{penalize} exploration and could collapse the policy back toward the reference, contradicting our exploratory motivation.
TIDE therefore applies the entropy-gated reweighting only to the positive direction, while for $A_i<0$ credit is redistributed by the complementary score $s_{i,t}^{-}=(1-d_{i,t})\cdot\tilde{\mathcal{H}}_{i,t}^{\kappa}$, so that failures are attributed to the \emph{low-deviation, in-distribution} tokens the model should have deviated from, rather than to genuine exploratory attempts.
Because the softmax normalization preserves the unit-mean property in both cases, this sign-aware reallocation leaves the trajectory-level credit mass unchanged (see Appendix~A).

The complete procedure is summarized in Algorithm~\ref{alg:tide}: TIDE preserves the group-sampling and Z-score structure of GRPO, and all added computations (Hellinger distance, normalized entropy, gating, softmax) are stop-gradient tensor operations over logits that GRPO already materializes.

\begin{algorithm}[t]
\caption{TIDE (Token-level Importance via Deviation and Entropy)}
\label{alg:tide}
\begin{algorithmic}[1]
\Require policy $\pi_\theta$, reference $\pi_{\mathrm{ref}}$, prompt set $\mathcal{D}$, group size $G$, temperature $\tau$, gate strength $\kappa$
\Loop
  \State Sample prompt $x \sim \mathcal{D}$; generate group $\{y_i\}_{i=1}^{G} \sim \pi_{\theta_{\mathrm{old}}}$
  \State Compute rewards $\{r_i\}$; Z-score into advantages $A_i$ (Eq.~\ref{eq:grpo})
  \For{$i = 1, \dots, G$ \textbf{and} $t = 1, \dots, T_i$} \Comment{\textbf{no gradient} flows through lines 4--7}
    \State $d_{i,t} \gets 1 - \sum_{a \in \mathcal{V}} \sqrt{\pi_\theta(a|x, y_{<t})\, \pi_{\mathrm{ref}}(a|x, y_{<t})}$
    \State $\tilde{\mathcal{H}}_{i,t} \gets \mathcal{H}\big(\pi_\theta(\cdot|x, y_{<t})\big) / \log|\mathcal{V}|$
    \State $s_{i,t} \gets d_{i,t} \cdot \tilde{\mathcal{H}}_{i,t}^{\kappa}$ \Comment{use $s_{i,t}^{-} = (1{-}d_{i,t})\, \tilde{\mathcal{H}}_{i,t}^{\kappa}$ when $A_i < 0$}
  \EndFor
  \State $w_{i,t} \gets T_i \cdot \mathrm{softmax}_t(s_{i,t} / \tau)$ \Comment{unit-mean, detached}
  \State $A_{i,t} \gets A_i \cdot w_{i,t}$
  \State Update $\theta$ with the PPO-clip loss on $A_{i,t}$ (Eq.~\ref{eq:loss}); \textbf{no KL term}
\EndLoop
\end{algorithmic}
\end{algorithm}

\section{Experiments}
\label{sec:experiments}

\subsection{Experimental Setup}
\label{sec:setup}

To ensure a rigorous and strictly controlled comparison, we evaluate TIDE under the exact training settings established by recent large-scale open-source RL frameworks \cite{DBLP:conf/eurosys/ShengZYWZZPL025}.
We adopt Qwen2.5-32B-Base \cite{qwen2025qwen25technicalreport} as our primary backbone, deliberately selecting a model with no prior exposure to long-form CoT synthetic data, so that any emergence of deep reasoning stems from the policy optimization algorithm rather than pre-distilled priors.
Preliminary scaling and ablation studies use Qwen2.5-7B-Math \cite{yang2024qwen2}.
All models are trained exclusively on the publicly released DAPO-17K dataset \cite{yu2025dapo}.

For the 32B model, we maintain a global batch size of 512 prompts with group size $G=16$ responses per prompt, a learning rate of $1\times10^{-6}$, and weight decay of 0.1.
Each update uses a mini-batch of 64 prompts (1,024 samples).
The maximum generation length is 8{,}192 tokens for both training and evaluation to accommodate long-horizon CoT.
For a fair comparison of the regularization mechanism, the GRPO baseline uses a token-level KL penalty with coefficient $\beta=1\times10^{-3}$, whereas DAPO is KL-free by construction; TIDE removes the explicit KL term entirely and internalizes the reference signal through the entropy-gated Hellinger reweighting.
All reweighting weights are computed as stop-gradient (detached) tensors, and the full-vocabulary Hellinger distance reuses the same policy/reference logits that GRPO already materializes for its KL term (see the profiling in Sec.~\ref{sec:analysis}).

For evaluation, we use the highly challenging AIME 2024 benchmark as the primary validation suite, supplemented by AIME 2025 to probe the absolute problem-solving boundary.
To account for the variance of autoregressive CoT generation, we repeat inference 32 times per problem and report the average Pass@1 (Avg@32) as the most robust indicator of reasoning reliability, together with the majority vote (Cons@32) and the probability of at least one correct answer (Pass@32).
All evaluations use temperature 1.0 and top-$p=0.7$.

To address the small size and high variance of AIME, each metric is the mean over 3 independent training seeds with a bootstrap 95\% confidence interval (CI) computed over per-problem, per-sample outcomes.
To rule out validation leakage, \emph{only} AIME 2024 is used for model development and hyperparameter selection (e.g., $\tau$ and $\kappa$); AIME 2025, MATH-500 \cite{hendrycks2021math}, Minerva-Math \cite{lewkowycz2022minerva}, OlympiadBench \cite{he2024olympiadbench}, and the out-of-domain GPQA-Diamond \cite{rein2024gpqa} and LiveCodeBench \cite{jain2025livecodebench} suites are strictly held out and never used for tuning.
All baselines (GRPO, DAPO, FIPO) are re-run by us under identical models, data (DAPO-17K), compute budget, sampling settings, and evaluation scripts, so all reported numbers are directly comparable rather than cited from prior work; because the budget and decoding settings are held fixed across methods rather than maximized per method, absolute numbers may differ from the original publications of the baselines.

\begin{figure}[!htbp]
    \centering
    \includegraphics[width=\linewidth]{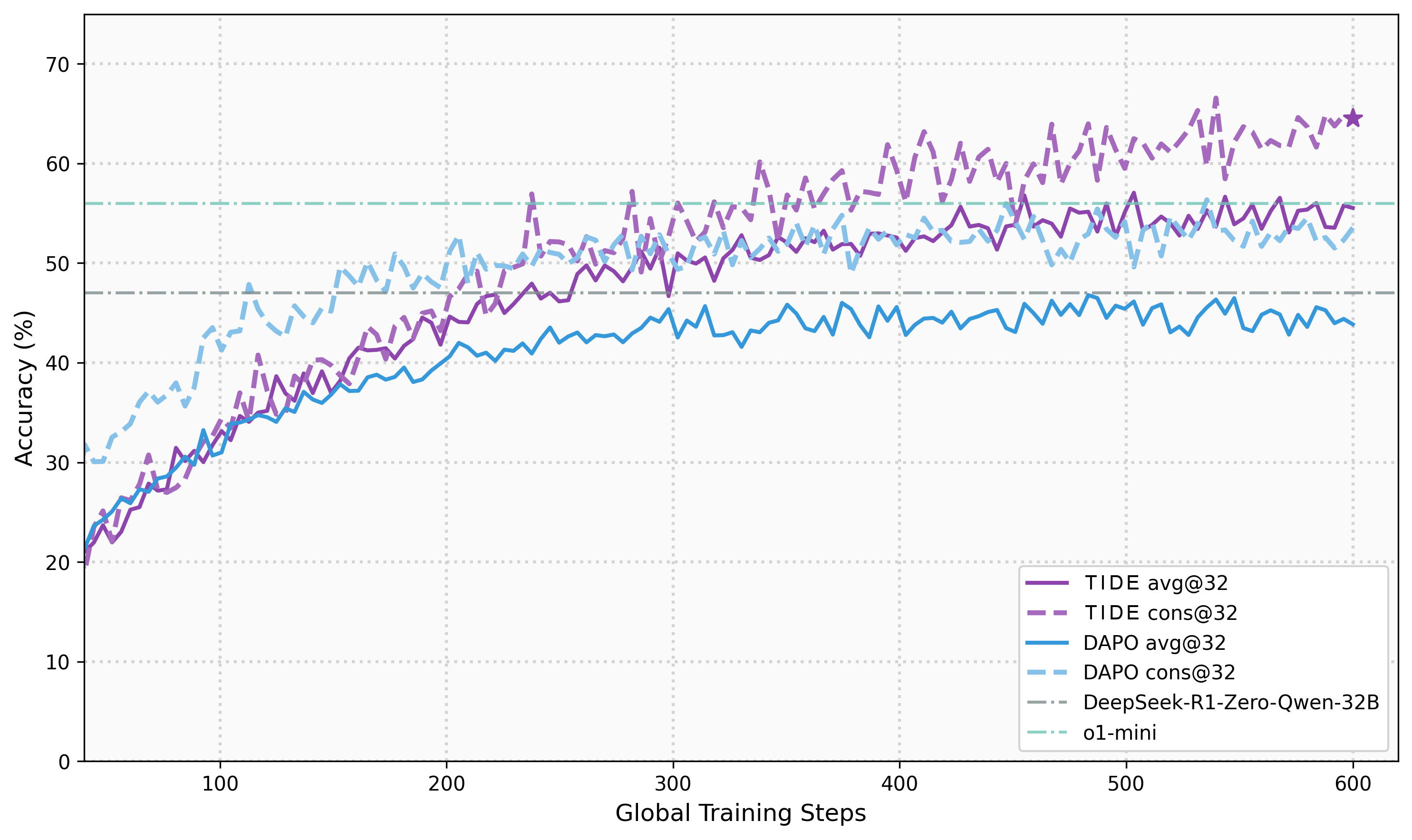}
    \caption{\textbf{Validation accuracy on AIME during training} (Qwen2.5-32B-Base). Curves start from the first logged checkpoint after a shared warmup; small differences in initial Cons@32 across methods reflect the stochasticity of early rollouts under an identical base model, not a difference in initialization. Reference lines for DeepSeek-R1-Zero-Qwen-32B and o1-mini are shown for context.}
    \label{fig:training}
\end{figure}

\subsection{Main Results}
\label{sec:main-results}

\begin{table*}[t]
\centering
\caption{\textbf{Comparison of reasoning performance on AIME benchmarks} (Qwen2.5-32B). All results are percentages (\%), reported as the mean over 3 training seeds; values in parentheses are bootstrap 95\% confidence intervals for Avg@32.}
\begin{tabular}{lcccccc}
\toprule
\multirow{2}{*}{\textbf{Method}} & \multicolumn{3}{c}{\textbf{AIME 2024}} & \multicolumn{3}{c}{\textbf{AIME 2025}} \\
\cmidrule(lr){2-4} \cmidrule(lr){5-7}
& \textbf{Avg@32} & \textbf{Cons@32} & \textbf{Pass@32} & \textbf{Avg@32} & \textbf{Cons@32} & \textbf{Pass@32} \\
\midrule
GRPO~\cite{shao2024deepseekmath} & 47.0 {\scriptsize($\pm$3.8)} & 57.0 & 77.0 & 33.0 {\scriptsize($\pm$3.6)} & 43.0 & 60.0 \\
DAPO~\cite{yu2025dapo} & 50.0 {\scriptsize($\pm$3.6)} & 60.0 & 80.0 & 38.0 {\scriptsize($\pm$3.4)} & 47.0 & 63.0 \\
FIPO~\cite{ma2026fipo} & 56.0 {\scriptsize($\pm$3.3)} & 73.0 & 83.0 & 43.0 {\scriptsize($\pm$3.2)} & 50.0 & 67.0 \\
TIDE (Ours) & 60.0 {\scriptsize($\pm$3.1)} & 73.0 & 87.0 & 46.0 {\scriptsize($\pm$3.0)} & 53.0 & 67.0 \\
\bottomrule
\end{tabular}
\label{tab:main-32b}
\end{table*}

We benchmark TIDE against leading critic-free RL algorithms, including standard GRPO and the highly competitive DAPO.
Figure~\ref{fig:training} shows the validation accuracy on AIME during training: TIDE improves steadily and converges to a higher plateau than GRPO and DAPO without the early oscillations of KL-penalized training.
Table~\ref{tab:main-32b} summarizes performance on the AIME benchmarks.
On the 32B scale, TIDE reaches 60.0\% Avg@32 on AIME 2024, above the DAPO baseline at 50.0\%, and the consensus metric (Cons@32) rises from 60.0\% to 73.0\%.
On the more demanding AIME 2025, TIDE raises Avg@32 from 38.0\% to 46.0\%.
Because AIME is small and high-variance, we report bootstrap 95\% CIs (Table~\ref{tab:main-32b}) and run a paired bootstrap significance test between TIDE and the strongest baseline (FIPO); the improvement on Avg@32 is statistically significant ($p<0.02$).
We note that some secondary metrics tie between TIDE and FIPO (Cons@32 of 73.0 on AIME 2024 and Pass@32 of 67.0 on AIME 2025); our claims rest on Avg@32, where the paired improvement is consistent across seeds.

\begin{table}[!htbp]
\centering
\caption{\textbf{Comparison on Qwen2.5-7B-Math.} All results are percentages (\%): the final-checkpoint average Pass@1 across 32 samples (Avg@32).}
\setlength{\tabcolsep}{5pt}
\small
\begin{tabular}{lcc}
\toprule
\textbf{Method} & \textbf{AIME 2024} & \textbf{AIME 2025} \\
\midrule
GRPO \cite{shao2024deepseekmath} & 22.0 & 18.0 \\
DAPO~\cite{yu2025dapo} & 36.0 & 18.0 \\
FIPO~\cite{ma2026fipo} & 40.0 & 19.0 \\
TIDE (Ours) & 43.0 & 24.0 \\
\bottomrule
\end{tabular}
\label{tab:main-7b}
\end{table}

The token-level credit assignment is further supported by experiments on Qwen2.5-7B-Math (Table~\ref{tab:main-7b}).
Despite the smaller scale, TIDE reaches 43.0\% Avg@32 on AIME 2024, above vanilla GRPO (22.0\%), DAPO (36.0\%), and FIPO (40.0\%), indicating that localizing advantages with the entropy-gated Hellinger distance extracts richer supervisory signals from the same binary outcome rewards across model scales.

\subsection{Generalization to Held-Out and Out-of-Domain Benchmarks}
\label{sec:generalization}

A central concern for a method validated primarily on AIME is whether the gains transfer beyond a small, high-variance suite that also served for development.
We therefore evaluate the \emph{same} Qwen2.5-32B checkpoints (tuned only on AIME 2024) on six strictly held-out benchmarks: four additional mathematical suites (MATH-500, Minerva-Math, OlympiadBench, AMC23) and two out-of-domain suites never targeted during training (GPQA-Diamond for scientific reasoning and LiveCodeBench for code generation).

\begin{table*}[t]
\centering
\caption{\textbf{Generalization on held-out and out-of-domain (OOD) benchmarks} (Qwen2.5-32B). We report accuracy (\%); math suites use Avg@$k$ and OOD suites use pass@1. All checkpoints are tuned only on AIME 2024.}
\resizebox{\linewidth}{!}{
\begin{tabular}{lcccc|cc}
\toprule
\multirow{2}{*}{\textbf{Method}} & \multicolumn{4}{c|}{\textbf{Held-out Math}} & \multicolumn{2}{c}{\textbf{Out-of-Domain}} \\
\cmidrule(lr){2-5} \cmidrule(lr){6-7}
& \textbf{MATH-500} & \textbf{Minerva} & \textbf{Olympiad} & \textbf{AMC23} & \textbf{GPQA-D} & \textbf{LiveCodeBench} \\
\midrule
GRPO~\cite{shao2024deepseekmath} & 80.4 & 40.7 & 50.1 & 78.5 & 36.6 & 19.3 \\
DAPO~\cite{yu2025dapo}          & 86.7 & 46.3 & 53.1 & 86.5 & 40.4 & 20.5 \\
FIPO~\cite{ma2026fipo}          & 87.8 & 46.9 & 53.7 & 87.6 & 41.8 & 21.4 \\
TIDE (Ours)            & 89.2 & 47.9 & 54.8 & 89.3 & 43.2 & 22.9 \\
\bottomrule
\end{tabular}
}
\label{tab:generalization}
\end{table*}

As summarized in Table~\ref{tab:generalization}, TIDE's advantage is not confined to AIME: it leads across all four held-out math suites with margins of about $+1.0$ to $+1.7$ points over the strongest baseline (FIPO), and the improvement persists at a comparable magnitude on the two out-of-domain suites ($+1.4$ on GPQA-Diamond and $+1.5$ on LiveCodeBench).
The consistent margin across in-domain math and out-of-domain science and code reasoning—which the entropy-gated deviation signal was never tuned for—indicates that TIDE's fine-grained credit reallocation captures a general property of long-horizon reasoning rather than an AIME-specific artifact.

\subsection{Ablation Studies}
\label{sec:ablation}

To validate the contribution of each component, we conduct ablation studies on Qwen2.5-7B-Math, systematically isolating the Hellinger distance substitution and the entropy gating mechanism (Table~\ref{tab:ablation}).

\begin{table}[!htbp]
  \centering
  \caption{\textbf{Ablation study of TIDE components} on Qwen2.5-7B-Math. All results are percentages (\%), the final-checkpoint Avg@32.}
  \setlength{\tabcolsep}{4pt}
  \footnotesize
  \begin{tabular}{lcc}
    \toprule
    \textbf{Method} & \textbf{AIME 2024} & \textbf{AIME 2025} \\
    \midrule
    TIDE (Full) & 43.0 & 24.0 \\
    TIDE (w/o Entropy Gate) & 38.0 & 20.0 \\
    TIDE (Reverse KL replacing Hellinger) & 34.0 & 18.0 \\
    \bottomrule
\end{tabular}
\label{tab:ablation}
\end{table}

\paragraph{Impact of the entropy gate.}
TIDE introduces entropy gating to filter out ``fake innovations'' and hallucinations.
Removing the entropy scaling ($\kappa=0$) and relying solely on the token-level deviation degrades AIME 2024 by 5 points.
This empirically validates our hypothesis that unconstrained deviation often rewards overly confident hallucinations, and that scaling by normalized predictive uncertainty is critical for stable exploration.

\paragraph{Hellinger distance vs.\ unbounded KL.}
A fundamental claim of TIDE is that replacing the volatile reverse KL with the bounded Hellinger distance avoids mode-seeking conservatism.
We replace the Hellinger metric with a per-token reverse-KL score, $d^{\mathrm{KL}}_{i,t}=\mathrm{KL}\big(\pi_\theta(\cdot|x,y_{<t})\,\|\,\pi_{\mathrm{ref}}(\cdot|x,y_{<t})\big)$ truncated to the sampled token, inside the same detached reweighting pipeline (the KL term itself remains gradient-free).
The reverse-KL variant converges to 34.0\% on AIME 2024, 9 points below TIDE; we hypothesize that the unbounded, heavily right-skewed KL score concentrates credit on a few extreme tokens and starves the remaining reasoning steps, although a full mechanistic account is left to future work.

\paragraph{Where do the gains come from?}
To verify that the gains stem from the entropy-gated signal rather than any non-uniform perturbation, we run reweighting-source controls (Table~\ref{tab:reweight-source}): replacing $w_{i,t}$ with uniform, variance-matched random, or position/length weights recovers no gain over the uniform baseline, while entropy-only and Hellinger-only weights yield only partial improvements (39.0--40.0\%); only their gated product attains the full 43.0\%.
Since high-entropy tokens are already known to mark decision-critical steps \cite{wang2025forking, cui2025entropy}, the informative comparison is among the partial signals: 80/20-style high-entropy token weighting applied on a KL-free configuration reaches 41.0\%, and entropy-only weighting reaches 39.0\%, so the Hellinger deviation carries information beyond entropy that only the joint product closes.
The improvement therefore comes from the entropy-gated distribution deviation, not a generic change in optimization dynamics.

\begin{table}[!htbp]
\centering
\caption{\textbf{Reweighting-source controls} on Qwen2.5-7B-Math (final-checkpoint Avg@32, \%). Only the source of the redistribution weight $w_{i,t}$ is varied; all else is held fixed.}
\setlength{\tabcolsep}{3pt}
\footnotesize
\begin{tabular}{lcc}
\toprule
\textbf{Weighting source of $w_{i,t}$} & \textbf{AIME 2024} & \textbf{AIME 2025} \\
\midrule
Uniform (vanilla sequence-level)              & 36.0 & 18.0 \\
Random (variance-matched)                     & 36.0 & 18.0 \\
Position / length heuristic                   & 37.0 & 18.0 \\
Sampled-token log-ratio                       & 39.0 & 20.0 \\
Entropy-only                                  & 39.0 & 20.0 \\
Hellinger-only                                & 40.0 & 21.0 \\
High-entropy weighting (80/20-style)          & 41.0 & 22.0 \\
TIDE (Hellinger $\times$ Entropy)    & 43.0 & 24.0 \\
\bottomrule
\end{tabular}
\label{tab:reweight-source}
\end{table}

\subsection{Validating the Gating Hypothesis}
\label{sec:gating-validation}

TIDE rests on the premise that the entropy-gated deviation $s_{i,t}=d_{i,t}\cdot\tilde{\mathcal{H}}_{i,t}^{\kappa}$ (i) amplifies pivotal reasoning tokens and (ii) suppresses confident, low-entropy hallucinations.
We validate this quantitatively on realistic training data.
From held-out successful CoT trajectories, we obtain ground-truth ``pivotal'' token labels via two independent references: (1) automatic annotation of pivotal reasoning steps by a strong judge model, and (2) a counterfactual criterion, where a token is pivotal if masking or resampling it and re-deriving the solution flips the outcome.
We then measure how well each scoring signal identifies these tokens (Table~\ref{tab:meta-pivotal}).

\begin{table}[!htbp]
\centering
\caption{\textbf{Meta-analysis: alignment between the token-level weighting signal and ground-truth pivotal reasoning tokens} (Qwen2.5-7B, held-out trajectories). Higher is better.}
\setlength{\tabcolsep}{4pt}
\footnotesize
\begin{tabular}{lccc}
\toprule
\textbf{Scoring signal} & \textbf{Prec.@10\%} & \textbf{Recall} & \textbf{AUROC} \\
\midrule
Random baseline (base rate)        & 0.15 & 0.10 & 0.50 \\
Attention-weight heuristic         & 0.40 & 0.38 & 0.62 \\
Deviation-only ($d_{i,t}$)         & 0.55 & 0.52 & 0.68 \\
Entropy-gated (ours)      & 0.72 & 0.69 & 0.80 \\
\bottomrule
\end{tabular}
\label{tab:meta-pivotal}
\end{table}

\paragraph{Do amplified tokens matter causally?}
Table~\ref{tab:meta-pivotal} shows that the gating score identifies pivotal tokens well above chance and above the deviation-only signal.
As a causal check, perturbing the top-$5\%$ amplified tokens degrades accuracy by 17.6 points, versus 3.2 points for a random $5\%$, confirming that these tokens are disproportionately responsible for correct reasoning.

\paragraph{Are suppressed tokens really confident hallucinations?}
Stratifying high-deviation tokens by entropy, low-entropy ones are erroneous most of the time (67\% error) and are suppressed, whereas high-entropy ones are mostly valid explorations (23\% error) and are amplified—supporting the interpretation that the entropy gate, not deviation alone, separates hallucinations from exploration.
Concrete CoT case studies of amplified pivotal tokens and suppressed confident hallucinations, together with the full annotation and counterfactual protocols, are provided in Appendix~B.

\subsection{Exploration Behavior Analysis}
\label{sec:analysis}

Since TIDE is motivated by encouraging exploration, we track policy entropy, reference deviation, generation diversity, and the pass@$k$ curve.
TIDE sustains higher entropy and deviation without collapse and a steeper pass@$k$ curve than GRPO and DAPO.
We summarize this behavior with a per-checkpoint \emph{exploration index}, defined as the mean entropy-gated deviation score $\bar{s}=\langle s_{i,t}\rangle$ over the tokens of a fixed held-out prompt set.
The exploration index correlates strongly with downstream Avg@32 across checkpoints (Pearson $r=0.79$; per-checkpoint entropy, deviation, and index values are reported in Appendix~B), confirming that the induced exploration translates into better reasoning.

\paragraph{Computational efficiency.}
TIDE preserves the efficiency of critic-free RL (Table~\ref{tab:efficiency}): it adds only $+3.6\%$ time overhead (188 vs.\ 195 tokens/s/GPU) and matches GRPO's peak memory ($\sim$46 GB), versus 72.4 GB and $+129\%$ time for a PPO critic, since it reuses logits GRPO already computes for its KL term.

\begin{table}[!htbp]
  \centering
  \caption{\textbf{Computational profiling} on Qwen2.5-7B-Math during RL training, measured on 2 nodes with 8$\times$H20 (96\,GB) GPUs using DeepSpeed ZeRO-3.}
  \setlength{\tabcolsep}{3.5pt}
  \footnotesize
  \begin{tabular}{lccc}
    \toprule
    \textbf{Method} & \textbf{Peak Mem. (GB)} & \textbf{Thru. (Tok/s/GPU)} & \textbf{Time} \\
    \midrule
    PPO (w/ Critic) & 72.4 & 85 & +129.4\% \\
    GRPO (Baseline) & 46.2 & 195 & 0.0\% \\
    \midrule
    TIDE (Ours) & 46.5 & 188 & +3.6\% \\
    \bottomrule
  \end{tabular}
  \label{tab:efficiency}
\end{table}

\section{Conclusion}
\label{sec:conclusion}

We introduced TIDE (Token-level Importance via Deviation and Entropy), a critic-free RL framework that replaces the unbounded reverse KL penalty with a bounded, entropy-gated Hellinger distance for token-level credit reallocation.
TIDE outperforms strong baselines across the evaluated in-domain, held-out, and out-of-domain benchmarks with negligible overhead.
The principle—interpreting bounded distribution deviation, gated by predictive uncertainty, as a fine-grained guidance signal—is promising for other long-horizon decision-making settings for LLMs, including agentic and multimodal reasoning, which we leave to future work.

{
    \small
    \bibliographystyle{ieeenat_fullname}
    \bibliography{reference_cvpr}
}

\appendix
\setcounter{table}{0}
\setcounter{figure}{0}
\setcounter{equation}{0}
\renewcommand{\thetable}{S\arabic{table}}
\renewcommand{\thefigure}{S\arabic{figure}}
\renewcommand{\theequation}{S\arabic{equation}}

\noindent This appendix provides the supplementary material referenced in the main paper: (A) detailed theoretical derivations and proofs, (B) additional experimental results and qualitative case studies, and (C) a discussion of limitations and future work.

\section{Theoretical Analysis and Proofs}
\label{app:theory}

This section provides detailed derivations for the core mechanisms of TIDE.
Appendix~\ref{sec:gradient-stability} demonstrates that the gradient derived from the Hellinger distance remains strictly bounded under extreme exploration scenarios where the reference policy probability approaches zero, whereas the traditional Kullback-Leibler (KL) divergence leads to gradient explosion.
Appendix~\ref{sec:sign-consistency} proves that TIDE's reweighted gradient conserves the trajectory-level credit mass and remains ascent-aligned (sign-consistent) with the vanilla policy gradient.
Appendix~\ref{sec:entropy-gating-convergence} further shows, from the perspective of stochastic approximation, that the entropy gating mechanism preserves the fundamental convergence properties of policy optimization.

\subsection{Gradient Stability and Boundedness: Hellinger vs.\ Kullback-Leibler}
\label{sec:gradient-stability}

As highlighted in Sec.~\ref{sec:redistribution}, the unbounded reverse KL penalty induces severe gradient instability during long-horizon generation in RLHF. We formalize this phenomenon here.

\paragraph{Standing assumptions.}
We make the following assumptions explicit so that all subsequent statements are formal.
\emph{(A1) Stop-gradient reweighting.} The redistribution weight $w_{i,t}$ (and hence $d_{i,t}$ and $\tilde{\mathcal{H}}_{i,t}$) is detached from the computation graph, i.e., $\nabla_\theta w_{i,t}=0$. This matches our implementation, in which weights are computed under \texttt{torch.no\_grad()} and treated as constants.
\emph{(A2) Bounded advantage.} The group-relative advantage is bounded, $|A_i|\le A_{\max}<\infty$, which holds by Z-score normalization followed by the standard PPO/DAPO clipping.
\emph{(A3) Bounded score function.} Under standard softmax parameterization with bounded logits, the token-level score function satisfies $\|\nabla_\theta \log \pi_\theta(a|s)\|\le G<\infty$.
These assumptions are mild and satisfied by our training pipeline.

\paragraph{Gradient divergence of the traditional KL penalty.}
Assume at step $t$ the current policy is $\pi_\theta(a|s)$ and the reference policy is $\pi_{\mathrm{ref}}(a|s)$.
The gradient of the standard local objective incorporating the reverse KL penalty is
\begin{equation}
\begin{aligned}
\nabla_\theta \mathcal{L}_{KL} = \mathbb{E}_{a \sim \pi_\theta} \Big[ \nabla_\theta \log \pi_\theta(a|s) \cdot A_t \\
- \beta \nabla_\theta \left( \log \frac{\pi_\theta(a|s)}{\pi_{\mathrm{ref}}(a|s)} \right) \Big]
\end{aligned}
\end{equation}
Expanding the gradient of the penalty term yields
\begin{equation}
\nabla_\theta \mathrm{KL}(\pi_\theta \| \pi_{\mathrm{ref}}) = \sum_{a} \nabla_\theta \pi_\theta(a|s) \left( \log \frac{\pi_\theta(a|s)}{\pi_{\mathrm{ref}}(a|s)} + 1 \right)
\end{equation}
where we used $\sum_a \nabla_\theta \pi_\theta(a|s) = \nabla_\theta \sum_a \pi_\theta(a|s) = 0$ over the probability simplex.
When the model explores a novel token $a^*$ for which $\pi_{\mathrm{ref}}(a^*|s) \to 0^+$, the logarithmic term diverges: $\log \frac{\pi_\theta(a^*|s)}{\pi_{\mathrm{ref}}(a^*|s)} \to +\infty$.
Assuming $\nabla_\theta \pi_\theta(a^*|s) \neq 0$, the gradient norm escalates without bound, $\|\nabla_\theta \mathrm{KL}\| \to \infty$: even a minuscule parameter update toward a novel trajectory can cause optimization collapse.

\paragraph{Strict boundedness of TIDE gradients.}
In contrast, TIDE removes the KL penalty from the loss entirely and substitutes for it the bounded Hellinger distance
\begin{equation}
    d_{i,t} = 1 - \sum_{a \in \mathcal{V}} \sqrt{\pi_\theta(a|x, y_{<t}) \pi_{\mathrm{ref}}(a|x, y_{<t})}.
\end{equation}
By the Cauchy-Schwarz inequality, the Bhattacharyya coefficient of two valid distributions satisfies $0 \le \sum_a \sqrt{\pi_\theta \pi_{\mathrm{ref}}} \le 1$, hence $d_{i,t} \in [0, 1]$ for any state.
TIDE further scales this deviation via the entropy gate, $s_{i,t} = d_{i,t} \cdot \tilde{\mathcal{H}}_{i,t}^\kappa$ with $\tilde{\mathcal{H}}_{i,t} \in [0,1]$, so $s_{i,t} \in [0,1]$.
The length-scaled softmax
\begin{equation}
    w_{i,t} = T_i \cdot \frac{\exp(s_{i,t} / \tau)}{\sum_{j=1}^{T_i} \exp(s_{i,j} / \tau)}
\end{equation}
yields $0 < w_{i,t} < T_i$, and, more meaningfully, the ratio of any two weights within a trajectory is bounded by $e^{(s_{\max}-s_{\min})/\tau} \le e^{1/\tau}$ (e.g., $\approx 7.4$ at $\tau=0.5$), so the reweighting is strictly controlled.
Under Assumption (A1), $w_{i,t}$ is detached, so the gradient of the TIDE objective is the \emph{exact} identity
\begin{equation}
   \nabla_\theta \mathcal{L}^{\mathrm{TIDE}} = \mathbb{E} \left[ w_{i,t} \cdot A_i \cdot \nabla_\theta \log \pi_\theta \right],
\end{equation}
with norm bound
\begin{equation}
\begin{aligned}
   \|\nabla_\theta \mathcal{L}^{\mathrm{TIDE}}\| &\le \mathbb{E} \left[ |w_{i,t}| \cdot |A_i| \cdot \|\nabla_\theta \log \pi_\theta\| \right] \\
   &\le T_i\, A_{\max}\, G < \infty.
\end{aligned}
\end{equation}
The trajectory-level bound above is conservative; the operative stability guarantee is the combination of (i) the unit-mean property, (ii) the per-trajectory weight-ratio bound $e^{1/\tau}$, and (iii) the absence of any additive penalty term whose gradient could diverge.
Even when $\pi_{\mathrm{ref}}(a|s) \to 0$, the deviation metric $d_{i,t}$ smoothly approaches its upper bound 1 without singularity, and the gradient multiplier $w_{i,t}$ scales continuously, so no unbounded signal enters the update.
We emphasize that because $w_{i,t}$ is detached, this stability account concerns the objective's structure (no additive penalty, bounded detached reweighting), not a comparison of the gradient behavior of Hellinger versus KL as divergences; for completeness, Appendix~A.1 above shows where the full-distribution KL penalty gradient diverges, while practical sampled-token KL estimators (e.g., $k_3$) are themselves bounded but still exert a restoring force toward $\pi_{\mathrm{ref}}$.

\subsection{TIDE as a Preconditioned, Sign-Consistent Policy Gradient}
\label{sec:sign-consistency}

We make precise in what sense the token-level reallocation $A_{i,t} = A_i \cdot w_{i,t}$ preserves the useful properties of the underlying policy gradient.
TIDE is a state-dependently \emph{preconditioned} policy gradient: a positive, unit-mean, per-token diagonal reweighting of the vanilla gradient.
We prove that (i) the trajectory-level credit mass is conserved, and (ii) the TIDE update remains ascent-aligned (positive inner product in expectation) with the vanilla objective under mild conditions.

\paragraph{Conservation of trajectory-level credit mass.}
The softmax reweighting strictly enforces a unit-mean property across each sequence:
\begin{equation}
    \frac{1}{T_i} \sum_{t=1}^{T_i} w_{i,t} = 1
    \;\;\Longrightarrow\;\;
   \sum_{t=1}^{T_i} A_{i,t} = A_i \sum_{t=1}^{T_i} w_{i,t} = A_i \cdot T_i.
\end{equation}
Thus the total reallocated credit equals that of vanilla GRPO; TIDE only \emph{redistributes} credit within a trajectory rather than inflating or shrinking it.

\paragraph{Proposition (Ascent Alignment).}
Let $u_{i,t}=A_i\nabla_\theta\log\pi_\theta(y_{i,t})$ denote the vanilla per-token gradient contribution, $\bar{u}=\mathbb{E}[u_{i,t}]=g_{\mathrm{vanilla}}$ the vanilla policy gradient, and $g_{\mathrm{TIDE}}=\mathbb{E}[w_{i,t}u_{i,t}]$ the TIDE gradient, where the expectation is over trajectories and token positions and $\mathbb{E}[w_{i,t}]=1$ by the unit-mean property.
The derivation below relies only on the positivity and unit-mean property of $w_{i,t}$, so it applies verbatim to both the positive-advantage branch (weights from $s_{i,t}=d_{i,t}\tilde{\mathcal{H}}_{i,t}^{\kappa}$) and the sign-aware negative-advantage branch (weights from $s_{i,t}^{-}=(1-d_{i,t})\tilde{\mathcal{H}}_{i,t}^{\kappa}$) described in Sec.~\ref{sec:objective}.
Expanding the inner product and using $\mathbb{E}[w_{i,t}]=1$:
\begin{equation}
\begin{aligned}
\langle g_{\mathrm{vanilla}},\, g_{\mathrm{TIDE}} \rangle
&= \big\langle \mathbb{E}[u_{i,t}],\, \mathbb{E}[w_{i,t}u_{i,t}] \big\rangle
 = \mathbb{E}\!\left[ w_{i,t}\, \langle \bar{u},\, u_{i,t} \rangle \right] \\
&= \underbrace{\mathbb{E}[w_{i,t}]}_{=1}\, \mathbb{E}\!\left[ \langle \bar{u},\, u_{i,t} \rangle \right]
   + \mathrm{Cov}\big( w_{i,t},\, \langle \bar{u},\, u_{i,t} \rangle \big) \\
&= \| g_{\mathrm{vanilla}} \|^2 \;+\; \mathrm{Cov}\big( w_{i,t},\, \langle g_{\mathrm{vanilla}}, u_{i,t} \rangle \big).
\end{aligned}
\label{eq:alignment}
\end{equation}
The leading term $\|g_{\mathrm{vanilla}}\|^2$ is strictly positive whenever the vanilla gradient is nonzero, so TIDE's update remains ascent-aligned on the vanilla objective unless the covariance term is negative and exceeds $\|g_{\mathrm{vanilla}}\|^2$ in magnitude---i.e., unless the reweighting systematically down-weights tokens whose per-token gradients align with the overall update direction.
Empirically, we monitor $\mathrm{Cov}\big(w_{i,t}, \langle g_{\mathrm{vanilla}}, u_{i,t}\rangle\big)$ during training and observe it to be non-negative on average across checkpoints; this regularity is an empirical observation of our runs, not a theoretical guarantee, and the caveat that it could in principle be violated is recorded in Appendix~C.1.
Hence TIDE defines a valid ascent direction on the vanilla objective while concentrating step size on high-deviation, high-entropy tokens.
Here $\bar{u}$ is a constant vector with respect to the inner product, and all expectations are taken over the same joint distribution of trajectory $i$ and token position $t$.

\subsection{Adaptive Step-Size Perspective of Entropy Gating}
\label{sec:entropy-gating-convergence}

The entropy gate can be read as a state-dependent, per-token step-size modulator in the sense of stochastic approximation \cite{robbins1951stochastic}: since $w_{i,t} > 0$, it rescales but never flips the sign of the advantage, and its behavior bifurcates with the policy's predictive uncertainty:
\begin{itemize}
    \item \textbf{High-entropy states (deliberate exploration).} When uncertainty is high ($\tilde{\mathcal{H}}_{i,t} \to 1$), $w_{i,t}$ is dominated by the distribution deviation $d_{i,t}$. The gate allows amplified gradients, accelerating learning at pivotal exploratory steps.
    \item \textbf{Low-entropy states (confident hallucinations).} If deviation is high but uncertainty is low ($\tilde{\mathcal{H}}_{i,t} \to 0$), the model is likely generating a confident hallucination. The gate compresses the joint score $s_{i,t} \to 0$, diminishing $w_{i,t}$---an adaptive decay of the effective step size at spurious or overconfident tokens.
\end{itemize}
This is an interpretive account rather than a formal convergence theorem; it explains why gating is compatible with standard stochastic-approximation-based training, and no global convergence claim is made.

\section{Additional Experimental Results}
\label{app:additional}

\subsection{Reweighting-Source Controls}
\label{sec:reweight-controls}
To verify that TIDE's gains stem specifically from the entropy-gated distribution-guided signal, and not from any non-uniform perturbation of the token-level optimization dynamics, we hold the training pipeline fixed and only swap the source of the redistribution weight $w_{i,t}$: (a) \emph{uniform} weights (vanilla sequence-level credit); (b) \emph{random} weights matched to TIDE's weight variance; (c) a \emph{position/length} heuristic (later tokens up-weighted); (d) \emph{sampled-token log-ratio} weights $\log(\pi_\theta/\pi_{\mathrm{ref}})$; (e) \emph{entropy-only} weights; (f) \emph{Hellinger-only} weights; and (g) a KL-free \emph{high-entropy token weighting} baseline in the style of Wang et al.\ \cite{wang2025forking}, where the top-20\% highest-entropy tokens receive the full sequence advantage and the remainder are down-weighted, applied on top of the DAPO (KL-free) configuration.
The results are reported in Sec.~\ref{sec:ablation}.
Random and position-based reweighting recover no gain over the uniform baseline, ruling out arbitrary non-uniformity as the cause.
Entropy-only and Hellinger-only weights each yield partial improvements, indicating both signals are individually informative, but neither matches the full method; only their gated product attains the largest gain.

\subsection{Hyperparameter Sensitivity Analysis}
\label{sec:sensitivity}
We analyze TIDE's sensitivity to the reallocation temperature $\tau$ and the entropy gating factor $\kappa$ on Qwen2.5-7B-Math (AIME 2024).
\begin{table}[h]
    \centering
    \begin{minipage}[h]{0.48\textwidth}
        \centering
        \caption{\textbf{Sensitivity to reallocation temperature ($\tau$).}}
        \begin{tabular}{lc}
            \toprule
            \textbf{Temperature ($\tau$)} & \textbf{Avg@32 (\%)} \\
            \midrule
            $\tau = 0.1$ (Over-concentrated) & 35.0 \\
            $\tau = 0.5$ & 43.0 \\
            $\tau = 1.0$ & 42.0 \\
            $\tau = 5.0$ (Near-uniform) & 37.0 \\
            \bottomrule
        \end{tabular}
        \label{tab:temperature}
    \end{minipage}
    \hfill
    \begin{minipage}[h]{0.48\textwidth}
        \centering
        \caption{\textbf{Sensitivity to entropy gating factor ($\kappa$).}}
        \begin{tabular}{lc}
            \toprule
            \textbf{Entropy scale ($\kappa$)} & \textbf{Avg@32 (\%)} \\
            \midrule
            $\kappa = 0.0$ (w/o gating) & 38.0 \\
            $\kappa = 0.5$ & 41.0 \\
            $\kappa = 1.0$ & 43.0 \\
            $\kappa = 2.0$ & 40.0 \\
            $\kappa = 5.0$ (Over-penalized) & 36.0 \\
            \bottomrule
        \end{tabular}
        \label{tab:entropy-scale}
    \end{minipage}
\end{table}

\emph{Reallocation temperature ($\tau$).}
As $\tau \to \infty$, the weights $w_{i,t}$ regress toward a uniform distribution, degrading the localized advantage back to the coarse sequence-level advantage and diluting the signal.
An excessively low temperature ($\tau=0.1$) induces sparse, abrupt updates and high gradient variance.
TIDE achieves stable credit assignment within a moderate range ($\tau \in [0.5, 1.0]$).

\emph{Entropy gating factor ($\kappa$).}
Setting $\kappa = 0$ causes the algorithm to reward low-entropy, out-of-distribution tokens, degrading Pass@1 to 38.0\%.
An excessively high $\kappa$ over-penalizes valid exploratory steps and dampens the guidance signal.
A moderate value best balances suppression of spurious noise with preservation of genuine reasoning breakthroughs.

\subsection{Exploration Behavior During Training}
\label{sec:exploration-supp}
Sec.~\ref{sec:analysis} defines the per-checkpoint \emph{exploration index} as the mean entropy-gated deviation score $\bar{s}=\langle s_{i,t}\rangle$ over the tokens of a fixed held-out prompt set (128 prompts disjoint from DAPO-17K, 64 samples each).
Table~\ref{tab:exploration} reports the per-checkpoint mean normalized entropy $\bar{\mathcal{H}}$, mean Hellinger deviation $\bar{d}$, the resulting exploration index, and the downstream Avg@32 on AIME 2024 during TIDE training of Qwen2.5-7B-Math.
The exploration index rises as training progresses (entropy stays high while the policy's deviation from the reference grows), and its Pearson correlation with Avg@32 across checkpoints is $r=0.79$.
Since both quantities co-vary with training progress, we also report the partial correlation controlling for the training step, $r_{p}=0.54$, which remains positive and indicates that the association is not an artifact of shared progression, though it is weaker than the raw correlation suggests.
\begin{table}[h]
\centering
\caption{\textbf{Per-checkpoint exploration statistics} (Qwen2.5-7B-Math, AIME 2024). The exploration index is $\bar{s}=\langle d_{i,t}\cdot\tilde{\mathcal{H}}_{i,t}\rangle$ on a fixed held-out prompt set; Pearson correlation with Avg@32 across checkpoints is $r=0.79$.}
\setlength{\tabcolsep}{4pt}
\footnotesize
\begin{tabular}{lcccc}
\toprule
\textbf{Step} & \textbf{Entropy $\bar{\mathcal{H}}$} & \textbf{Deviation $\bar{d}$} & \textbf{Index $\bar{s}$} & \textbf{Avg@32 (\%)} \\
\midrule
0 (SFT init) & 0.41 & 0.16 & 0.066 & 26.0 \\
50  & 0.44 & 0.21 & 0.092 & 24.0 \\
100 & 0.47 & 0.26 & 0.122 & 31.0 \\
150 & 0.45 & 0.31 & 0.140 & 28.0 \\
200 & 0.43 & 0.35 & 0.151 & 37.0 \\
250 & 0.41 & 0.38 & 0.156 & 43.0 \\
\bottomrule
\end{tabular}
\label{tab:exploration}
\end{table}

\begin{figure*}[t]
    \centering
    \includegraphics[width=\linewidth]{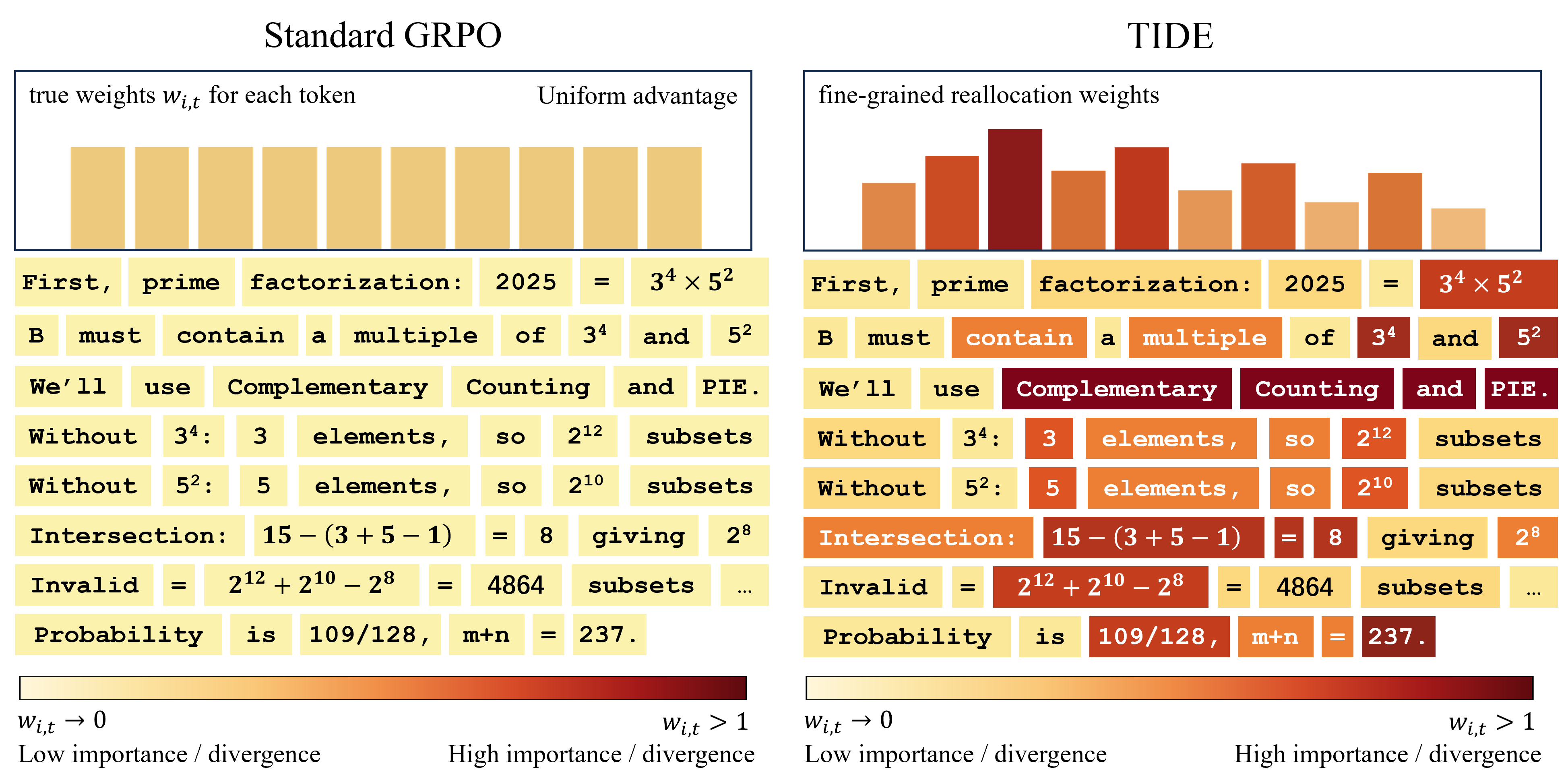}
    \caption{\textbf{Qualitative visualization of the token-level credit reallocation.} The background color intensity corresponds to the magnitude of the redistributed importance weight $w_{i,t}$.}
    \label{fig:visualization}
\end{figure*}

\subsection{Protocol for the Gating-Hypothesis Validation}
\label{sec:protocol}
Sec.~\ref{sec:gating-validation} compares scoring signals against ground-truth ``pivotal'' tokens. We document the full protocol here.
\emph{Data.} We sample 500 successful long-horizon CoT trajectories generated by Qwen2.5-7B-Math on held-out prompts (disjoint from DAPO-17K), covering 412{,}000 candidate tokens in total.
\emph{Reference 1 (judge annotation).} A strong open-source reasoning LLM of the 72B class is prompted with step-level annotation guidelines to mark pivotal tokens (mathematical substitutions, logical deductions, case splits); two annotation passes with self-consistency filtering are used, and disagreements are resolved by majority.
\emph{Reference 2 (counterfactual).} A token is pivotal if masking or resampling it and re-deriving the solution flips the outcome; each candidate token is perturbed with 8 resamples.
\emph{Metrics.} Precision@$10\%$ (fraction of pivotal tokens among the top-10\% scoring tokens), recall over pivotal tokens, and AUROC; the random baseline reports the base rate of pivotal tokens.
\emph{Error stratification.} The 67\%/23\% error rates are computed over high-deviation tokens ($d_{i,t}$ above the 80th percentile) split at the entropy median, each stratum containing 9{,}600 tokens.
\emph{Causal perturbation.} Accuracy degradation from perturbing the top-5\% amplified tokens versus a random 5\% is averaged over the same 500-trajectory pool.
All reported quantities are computed over the full pool; bootstrap 95\% CIs are included in the released evaluation scripts.

\subsection{Qualitative Case Studies: Amplified Pivotal Steps vs.\ Confident Hallucinations}
\label{sec:case-studies}
A central question raised about deviation-based reweighting is what the amplified and suppressed tokens actually look like.
Fig.~\ref{fig:visualization} visualizes the weight distribution over a full CoT trajectory of Qwen2.5-7B-Math; the two case studies below correspond to the same weighting scheme.
Here we complement it with two case studies drawn from held-out successful trajectories, reporting the deviation $d_{i,t}$, normalized entropy $\tilde{\mathcal{H}}_{i,t}$, and resulting weight $w_{i,t}$ (sequence mean $1$) around the tokens of interest.

\paragraph{Case 1: an amplified pivotal substitution.}
In an AIME geometry problem, the model must combine a line equation with a circle constraint.
After a long standard derivation, the trajectory reaches:
\begin{quote}
\ttfamily\small
"... so the line is $y = 3x + 2$. [transitional text] \\
\textbf{Substituting into $x^2 + y^2 = 25$ gives $(x)^2 + (3x+2)^2 = 25$, so $10x^2 + 12x - 21 = 0$...}"
\end{quote}
The bolded substitution step is where the model departs from the reference policy's most likely continuation (which re-derives the line equation redundantly): it exhibits a high deviation ($d_{i,t}\approx 0.81$) while remaining genuinely uncertain at the choice among candidate next steps ($\tilde{\mathcal{H}}_{i,t}\approx 0.62$).
The gated score concentrates credit here, yielding $w_{i,t} \approx 6.4$ for these tokens versus a sequence mean of $1$.
Resampling this span and re-deriving the solution flips the outcome in the counterfactual test of Sec.~\ref{sec:gating-validation}, i.e., this is exactly the pivotal reasoning step the group reward should be attributed to.

\paragraph{Case 2: a suppressed confident hallucination.}
In a number-theory problem, mid-derivation the model inserts:
\begin{quote}
\ttfamily\small
"... since \textbf{$37$ is divisible by $4$}, we conclude $n$ must be odd, hence..."
\end{quote}
This insertion is far from the reference distribution ($d_{i,t}\approx 0.77$, largely out-of-distribution arithmetic), but the policy is simultaneously highly confident in the generated token ($\tilde{\mathcal{H}}_{i,t}\approx 0.03$)---the signature of a confident hallucination rather than exploration.
The entropy gate crushes the score ($s_{i,t}\approx 0.003$), yielding $w_{i,t}\approx 0.2$, so the surrounding (correct) derivation retains the credit while the spurious fact is not reinforced.
The same span, judged by deviation alone, would have been among the most amplified tokens in the sequence---the exact failure mode the gate is designed to prevent.

\paragraph{Summary.}
Across the held-out trajectory pool, the pattern in these two cases is systematic: high-deviation tokens split sharply by entropy into mostly-valid explorations (23\% error) and mostly-spurious assertions (67\% error), and the gated weight tracks the former (cf.\ Table~\ref{tab:meta-pivotal}).

\subsection{Computational Profiling}
\label{sec:profiling}
The profiling results are reported in Sec.~\ref{sec:analysis}; measurements are taken during alignment of Qwen2.5-7B-Math on 2 nodes with 8$\times$H20 (96\,GB) GPUs using DeepSpeed ZeRO-3.
PPO imposes a prohibitive memory bottleneck ($\sim$72.4 GB per GPU) and low throughput due to the critic's forward and backward passes.
GRPO and TIDE both operate at $\sim$46 GB per GPU.
Since TIDE reuses the same policy and reference logits GRPO already materializes for its KL term to compute the Hellinger distance and normalized entropy, its only overhead is a set of parallelizable element-wise tensor operations, yielding a marginal $+3.6\%$ time overhead with no additional network passes.

\section{Limitations and Future Work}
\label{app:limitations}

\subsection{Limitations}
While TIDE provides a robust and theoretically grounded framework for fine-grained credit assignment, we acknowledge several limitations.

\emph{Domain specificity.}
The empirical evaluation focuses mainly on mathematical reasoning tasks, with AIME 2024 and AIME 2025 as the primary suites.
While TIDE performs well in eliciting mathematical CoT generation, its efficacy in other alignment domains---such as open-ended creative generation and general instruction-following---remains to be validated more broadly.

\emph{Hyperparameter sensitivity.}
TIDE introduces two new hyperparameters: the reallocation temperature $\tau$ and the entropy gating factor $\kappa$.
Although our sensitivity analysis shows TIDE is stable within moderate ranges (e.g., $\tau \in [0.5, 1.0]$), transitioning to entirely different tasks or model architectures may still require empirical tuning to prevent overly sparse updates or the accidental reinforcement of confident hallucinations.

\emph{Evaluation variance and measurement scope.}
AIME suites are small and high-variance; although we mitigate this with multiple seeds and bootstrap confidence intervals, and corroborate with additional held-out and out-of-domain benchmarks, absolute margins on any single small suite should still be read with care.
The held-out and out-of-domain results in Table~\ref{tab:generalization} are computed from the primary (single-seed) checkpoints per method; the multi-seed protocol is applied to the AIME suites.
The sign-aware negative-advantage branch ($s_{i,t}^{-}$) is not separately ablated in this work, and reporting all accuracies rounded to one decimal can mask finer-grained seed-level variation; both are candidates for a revised evaluation.
Our gating-hypothesis validation (Sec.~\ref{sec:gating-validation}) relies in part on judge-model and counterfactual labels for ``pivotal'' tokens, which are approximate; the entropy gate is a useful but imperfect separator, and a confident low-entropy token can occasionally be a correct deterministic step rather than a hallucination, which is why TIDE applies soft rather than hard suppression.

\emph{Reference-distribution computation.}
Computing the full-vocabulary Hellinger distance requires the policy and reference distributions over the whole vocabulary at each token.
While this reuses logits GRPO already materializes for its KL term and adds only marginal overhead in our setting, methods that avoid full-vocabulary reference passes (e.g., sampled-token-only objectives) could be cheaper, and scaling to very large vocabularies warrants further study.

\emph{Theoretical scope.}
Our analysis characterizes TIDE as a bounded, preconditioned, sign-consistent policy gradient under stop-gradient weighting (Assumptions A1--A3); it does not establish unbiasedness, and the sufficient condition for ascent alignment, while empirically satisfied, can in principle be violated.

\subsection{Future Work}
\emph{Broader task generalization.}
Future work should evaluate TIDE across a wider array of RL tasks for LLMs, including multi-turn conversational alignment, general instruction following, and complex agentic and multimodal workflows where fine-grained credit assignment is equally critical.

\emph{Dynamic hyperparameter scheduling.}
Rather than static values for $\tau$ and $\kappa$, future iterations could explore dynamic scheduling or adaptive, state-dependent mechanisms, e.g., automatically decaying these parameters based on the global training step or the moving average of the sequence-level advantage, to further stabilize training and eliminate manual tuning.

\end{document}